\documentclass[lettersize,journal]{IEEEtran}
\usepackage{amsmath,amsfonts}
\usepackage{array}
\usepackage[caption=false,font=normalsize,labelfont=sf,textfont=sf]{subfig}
\usepackage{textcomp}
\usepackage{stfloats}
\usepackage{url}
\usepackage{verbatim}
\usepackage{graphicx}
\usepackage{cite}
\usepackage[colorlinks=true]{hyperref}
\usepackage[ruled,linesnumbered]{algorithm2e}
\usepackage{multirow}
\usepackage{tikz}
\usepackage{booktabs}
\usepackage{graphicx}
\usepackage{booktabs}

\usepackage{booktabs}
\usepackage[switch]{lineno}
\begin{document}

\title{Robust Noisy Label Learning via Two-Stream Sample Distillation}

\author{Sihan Bai, 
       Sanping Zhou,~\IEEEmembership{Member,~IEEE},
       Zheng Qin,
       Le Wang,~\IEEEmembership{Senior Member,~IEEE},
       Nanning Zheng,~\IEEEmembership{Fellow,~IEEE}
\thanks{
This work was supported in part by NSFC under Grants 62088102, 12326608 and 62106192, Natural Science Foundation of Shaanxi Province under Grant 2022JC-41, and Fundamental Research Funds for the Central Universities under Grant XTR042021005.~({\it Corresponding author: Sanping Zhou, E-mail: spzhou@mail.xjtu.edu.cn.})}
\thanks{
Sihan Bai, Sanping Zhou, Le Wang and Nanning Zheng are all with the National Key Laboratory of Human-Machine Hybrid Augmented Intelligence, National Engineering Research Center for Visual Information and Applications, and Institute of Artificial Intelligence and Robotics, Xi'an Jiaotong University, Shaanxi 710049, China.}
}

\markboth{IEEE TRANSACTIONS ON MULTIMEDIA}%
{Shell \MakeLowercase{\textit{et al.}}: A Sample Article Using IEEEtran.cls for IEEE Journals}


\maketitle

\begin{abstract}
Noisy label learning aims to learn robust networks under the supervision of noisy labels, which plays a critical role in deep learning. Existing work either conducts s
ample selection or label correction to deal with noisy labels during the model training process. In this paper, we design a simple yet effective sample selection framework, termed Two-Stream Sample Distillation~(TSSD), for noisy label learning, which can extract more high-quality samples with clean labels to improve the robustness of network training. Firstly, a novel Parallel Sample Division~(PSD) module is designed to generate a \emph{certain} training set with sufficient reliable positive and negative samples by jointly considering the sample structure in feature space and the human prior in loss space. Secondly, a novel Meta Sample Purification~(MSP) module is further designed to mine adequate semi-hard samples from the remaining \emph{uncertain} training set by learning a strong meta classifier with extra golden data. As a result, more and more high-quality samples will be distilled from the noisy training set to train networks robustly in every iteration. Extensive experiments on four benchmark datasets, including CIFAR-10, CIFAR-100, Tiny-ImageNet, and Clothing-1M, show that our method has achieved state-of-the-art results over its competitors.
\end{abstract}


\section{Introduction}
\IEEEPARstart{T}{he} significant achievement of deep learning can be attributed primarily to Deep Neural Network (DNN) training using a large-scale dataset with human-annotated labels~\cite{Zhou_Wang_Wang:2017,He_Zhang_Ren:2016,Zhou_Wang_Wang:2023}. However, the process of labeling large amounts of data with high-quality annotations is labor intensive and time-consuming. To address this problem, researchers have extensively studied the Noisy Label Learning~(NLL) problem~\cite{10.5555/3327757.3327944,Song_Kim_Park:2022}, which focuses on how to train robust networks using a large number of samples with noisy labels. 

In general, existing work adopts either the sample selection paradigm~\cite{Shu_Xie_Yi:2019,10.5555/3305381.3305406,10.5555/3327757.3327944} or the label correction paradigm~\cite{Zheng_Awadallah_Dumais:2021,Feng_2022_BMVC} to address the NLL problem, both of which expect to involve more samples with clean labels in the training process. What is different, the former tries to choose samples with clean labels, while the latter aims at correcting the wrong labels of samples. Because label correction methods can potentially take more samples in network training, they can obtain better results than sample selection methods in close dataset evaluation. However, regarding open environment scenarios, there are sufficient samples with the correct labels to train networks. Therefore, the sample selection paradigm is more applicable due to its simplicity in application. As a result, more and more attentions~\cite{li2020dividemix,ortego2021multiobjective,ICML2019_UnsupervisedLabelNoise} have been paid to how to distill more training samples with clean labels to solve the NLL problem.

 \begin{figure}[t]
\centering
\includegraphics[scale=0.4]{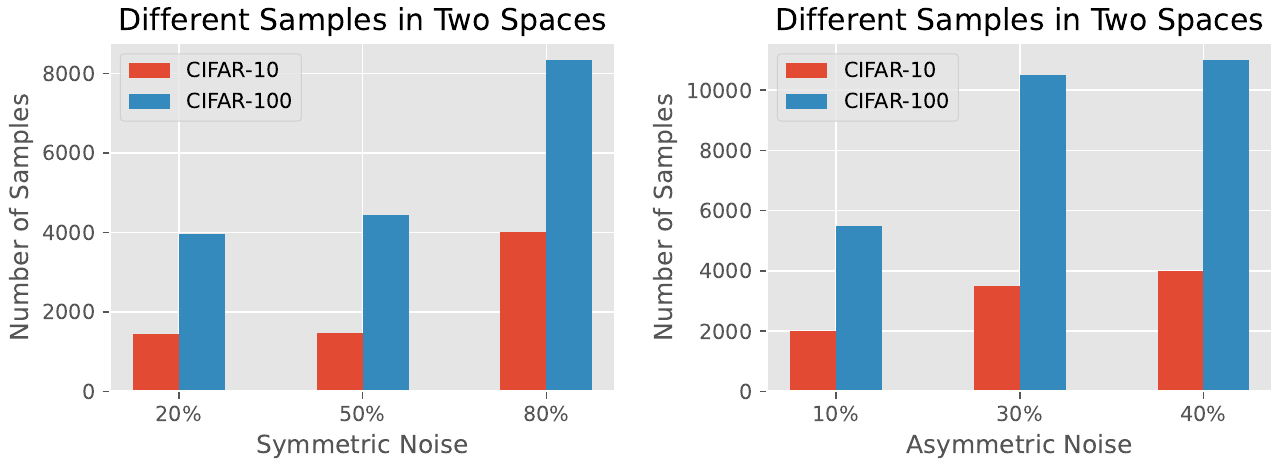}
\caption{Statistics on the number of inconsistent sample selection between loss space and feature space. 
Under different noise and different datasets, there are always inconsistent parts of the data filtered using the loss and feature method, reflecting the differences between the two methods. The experimental results are obtained by using the model trained in the first epoch after the warm-up training on the CIFAR-10 and CIFAR-100 datasets.}
\label{label_mot}
\end{figure}

The critical issue of sample selection lies in how to judge the reliability of noisy labels in the training process. To address this problem, both the small loss criterion~\cite{ijcai2021p340,10.5555/3327757.3327944,9568980,9552480} and feature clustering~\cite{wu2020topological, pmlr-v119-bahri20a, pmlr-v80-bao18a, 9576627} methods have been extensively explored in recent years. For example, MentorNet~\cite{jiang2018mentornet} uses a data-driven curriculum learning regime to involve high-confidence samples from easy to hard, while SSR~\cite{Feng_2022_BMVC} applies a sample selection algorithm based on a KNN classifier to select more training samples with clean labels. To our best understanding, the two methods take different mechanisms to select reliable samples. In particular,~\textbf{(1)} the former loss-based methods mainly embed human prior because noisy labels are usually provided by humans in practice;~\textbf{(2)} the latter feature-based methods mainly explore sample structure because sample similarity remains a critical clue in clustering algorithms.  As shown in Fig.~\ref{label_mot}, the presence of this difference is evident; particularly with 80\% noise, these inconsistent samples may represent 20\% of the total dataset comprising 50,000 samples. 
The two methods focus on different features; loss space pays more attention to human prior dependent features, while feature space concerns features of the data itself, so the combination of the two allows for a more comprehensive assessment of the sample, which is beneficial for mining semi-hard samples within it.

\begin{figure}[t]
\centering
\includegraphics[scale=0.9]{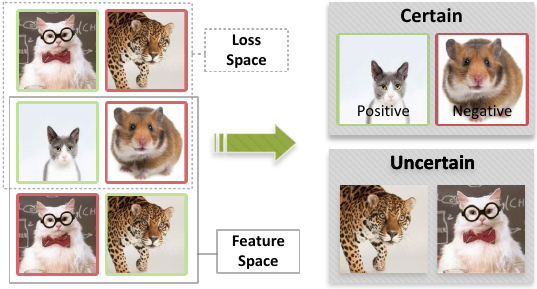}
\caption{Illustration of PSD module, in which it divides the training samples into one certain set and another uncertain set based on the information in both feature space and loss space. Samples with green and red borders are considered clean label and noise label samples, respectively.
}
\label{mot1}
\end{figure}

In this paper, we design a novel Two-Stream Sample Distillation~(TSSD) framework to train a strong network under the supervision of noisy labels. Specifically, it consists mainly of a Parallel Sample Division~(PSD) module and another Meta Sample Purification~(MSP) module, in which the former performs the training data partitioning by jointly considering the screening results in either feature space or loss space, while the latter conducts the semi-hard sample mining by learning a meta classifier with extra golden data. As shown in Fig.~\ref{mot1}, the PSD module partitions the training samples into two disjoint sets, \emph{i.e.}, \emph{certain} set and \emph{uncertain} set, in which:~\textbf{(1)} the \emph{certain} set mainly includes the positive and negative samples which are accepted as clean samples and rejected as noisy samples with high confidence, respectively; ~\textbf{(2)} the \emph{uncertain} set mainly includes these semi-hard samples which cannot be judged as clean or noisy ones due to their low confidence in both feature space and loss space. To further involve more semi-hard samples in network training, the MSP module takes both positive and negative samples in the \emph{certain} set as golden data, to learn a binary classifier that can verify whether these samples in the \emph{uncertain} set can be voted into the positive set. As a result, more and more high-quality samples can be distilled from the entire training data, which can consistently learn a robust network through iteration. We conducted extensive experiments to evaluate our method from different points of view, which have achieved state-of-the-art performance with different noise types and noise rates on CIFAR-10, CIFAR-100, and Tiny-ImageNet, as well as the real-world noisy dataset Clothing-1M.

The main contributions of this work can be highlighted as follows:
\begin{itemize}
    \item We design a novel Two-Stream Sample Distillation method for robust noisy label learning, which can mine more and more high-quality samples with clean labels to train networks.
    \item We design a novel Parallel Sample Division module for reliable data partition, which can jointly consider the sample structure in feature space and the human prior in loss space.
    \item We design a novel Meta Sample Purification module for semi-hard sample mining, which can learn a meta classifier to refind more positive samples from the \emph{uncertain} set to \emph{centrain} set.
\end{itemize}

The remainder of this paper is organized as follows: Section~\ref{sec:relatework} discusses related work. Section~\ref{sec:method} presents the technical details of the proposed TSSD. The experimental results and discussion are presented in Section~\ref{sec:experiments}. Ablation studies are shown in Section~\ref{sec:ablation}. Finally, we conclude the paper in Section~\ref{sec:conclusion}.

\section{Related Work}
\label{sec:relatework}
Many recent works have adopted different sample selection technologies to deal with the noisy label learning problem, which could be simply divided into two categories, \emph{i.e.}, the loss-based method and the feature-based method. These methods are reviewed in the following paragraphs.

\textbf{Loss-based Methods.} According to the study on the memorization effect~\cite{10.5555/3305381.3305406}, DNNs initially learn simple patterns and then gradually memorize all samples. As a result, a large number of previous works~\cite{10.5555/3327757.3327944,jiang2018mentornet,shen2019learning,9732177,10167857} treat the samples with small training loss as clean samples and then took them to train DNNs in a supervised manner. The main challenge of these methods lies in setting a suitable threshold to determine which samples are easy enough. To address this problem, the meta-learning regime has been applied to learn an adaptive weighting scheme, in which samples with clean labels will be given large weights to participate in model training. Typical methods include Meta-Weight-Net~\cite{Shu_Xie_Yi:2019}, MetaCleaner~\cite{8953770} and Meta Label Correction~\cite{Zheng_Awadallah_Dumais:2021}, in which they all learn a sample weighting function by using a small portion of labeled samples as meta-data. In recent years, the semi-supervised learning framework~\cite{10.5555/3454287.3454741, 10.5555/3495724.3495775} has been widely applied to solve the noisy label learning problem. This line of methods typically starts by selecting a clean label set and a noisy label set based on the small-loss criterion, which are then used for semi-supervised training as labeled and unlabeled data, respectively. Typical methods include the well-known DivideMix~\cite{li2020dividemix} and U{\footnotesize{NI}}C{\footnotesize{ON}}~\cite{Karim_Rizve_Rahnavard:2022}, in which the former divides the dataset by the loss distribution of each sample with a GMM, while the latter designs an adaptive sample selection scheme for the JS-divergence distribution of samples with the same label. One limitation of these methods is that they use only the information in loss space but ignore the information in feature space to conduct sample selection, making it difficult to keep the quality of samples in the clean set.

\begin{figure*}[ht]
\centering
\includegraphics[scale=0.65]{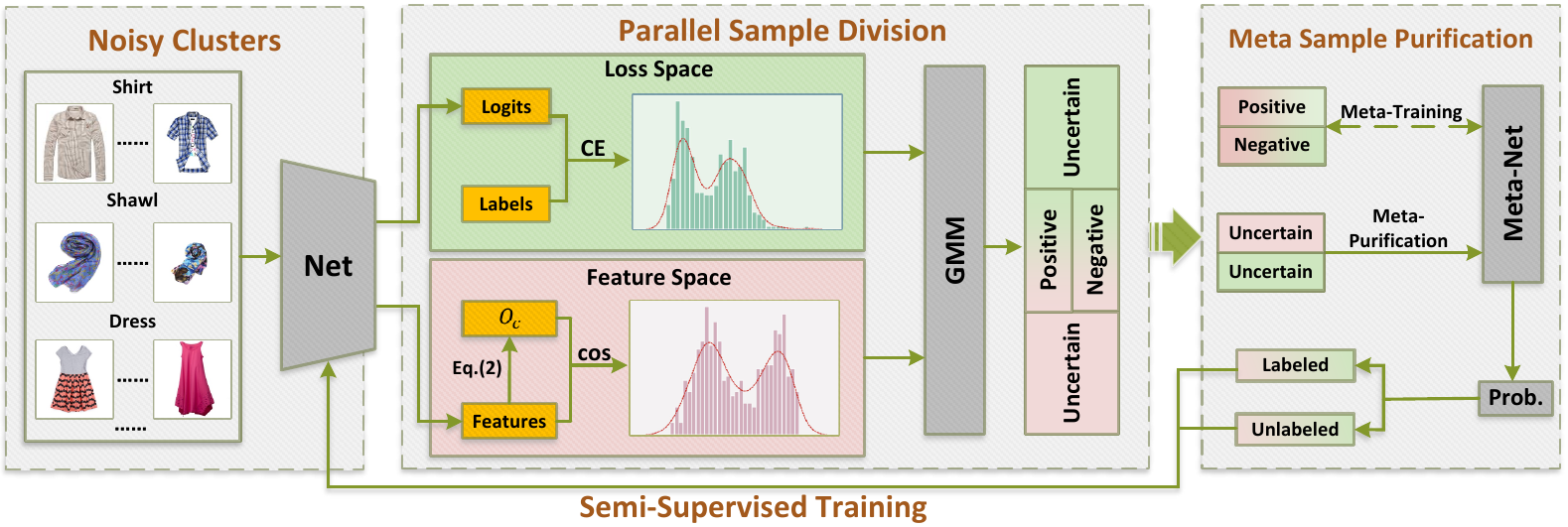}
\caption{Framework of our Two-Stream Sample Distillation. First, the training samples are divided into different noisy clusters based on their given labels. Second, the backbone extracts feature from the samples for each cluster, which are then passed on to the subsequent modules:~\textbf{(1)} The PSD module jointly considers the human prior in loss space and the sample structure in feature space, so as to generate the \emph{positive} and \emph{negative} sample in the \emph{centain} set; \textbf{(2)} The MSP module trains a binary classifier with golden data to further identify additional semi-hard samples in the \emph{uncertain} set. Third, we take samples in the \emph{certain} set as labeled data and samples in the \emph{uncertain} set as unlabeled data, after which an off-the-shelf semi-supervised learning algorithm is taken to train a robust network.}
\label{pipeline}
\end{figure*}

\textbf{Feature-based Methods.} Based on the assumption that samples with the same labels will have a similar appearance, feature-based methods often apply different clustering algorithms~\cite{10.1145/3136625,10.1145/1835804.1835848,9710910,10.1007/s00454-006-1271-x,9645225} to estimate pseudo-labels of samples in the training process. As sample similarity is a crucial factor in clustering algorithms, this line of methods focuses mainly on exploring the structure of samples. For example, several works have applied the KNN clustering algorithm to find possible samples with clean labels, including RkNN~\cite{gao2018resistance}, Deep KNN~\cite{pmlr-v119-bahri20a}, SSR~\cite{Feng_2022_BMVC}, TopoFilter~\cite{wu2020topological}, GLMNN-PLL~\cite{9529072}. In addition, some methods focus on how to cluster samples into different groups using the GMM model. For example, GMDA~\cite{liu2022gmm} believes that the probability distribution of each class in the dataset is not a single Gaussian distribution; instead, it should be treated as a mixture of Gaussian distribution. Furthermore, TCL~\cite{10203735} and CC~\cite{zhao2022centrality} model the data distribution through a GMM and detect samples with incorrect labels as those out-of-distribution ones. What is different, there are also a lot of methods~\cite{Northcutt2017LearningWC,Wang2019LessIB,10.1007/s11063-018-9963-9} conduct sample selection without using clustering. For example, Rank Pruning~\cite{Northcutt2017LearningWC} presents a method of confidence learning, in which it first estimates the noise rates of samples and then removes the least confident samples based on the resulting noise rates. In addition, Less Is Better~\cite{Wang2019LessIB} employs an influence function to estimate the impact of each training sample; therefore, more reliable samples with clean labels can be selected to train networks. One limitation of these methods is that they only use the information in feature space but ignore the information in loss space to conduct sample selection, making it hard to keep the quality of samples in the clean set.

What is different, our TSSD method attempts to address the respective issues of sample selection by jointly considering the human prior in loss space and the sample structure in feature space. As a result, the complementarity between feature space and loss space will be fully utilized to mine more and more high-quality samples with clean labels to train networks.

\section{Proposed Method}\label{sec:method}
\subsection{Preliminaries}
We propose a simple yet effective framework called Two-Stream Sample Distillation~(TSSD) for learning with noisy labels. As shown in Fig.~\ref{pipeline}, our algorithm consists of two modules: \textbf{(1)} Parallel Sample Division~(PSD) and \textbf{(2)} Meta Sample Purification~(MSP). In particular, the PSD module partitions the noisy labeled dataset by considering both the screening results in feature space and loss space; while the MSP module further identifies additional samples in the \emph{uncertain} set by training a robust binary classifier with golden data. 
Finally, we employ an off-the-shelf semi-supervised learning algorithm for a robust network based on the samples in the \emph{certain} set.

In the noisy label learning problem, we often encounter a training dataset with noisy labels. In the context of a classification task involving $K$ classes and $N$ images, 
the set of sample labels can be denoted as $\mathcal{{K}} = \{k_i\}_{i=1}^K$. 
The dataset can be denoted as $\mathcal{{D}} = \{(\mathcal{X},\mathcal{\tilde{Y}})\} = \{(x_n,\tilde{y}_{   n})\}_{n=1}^N$, where $\mathcal{X}$ denotes the image set and $\mathcal{\tilde{Y}}$ denotes the corresponding label set. We instantiate the DNN model with a CNN backbone,  $f(\cdot;\theta)$; a projection head, $h(\cdot;\psi)$; a classifier $g(\cdot;\phi)$. Based on these settings, we divide the training data into one clean-labeled set and another noisy-labeled set, in which the former set contains samples with clean labels, while the latter set contains the samples with noisy labels. Finally, we use an off-the-shelf semi-supervised learning regime to train a robust network, in which samples with clean labels are taken as the labeled set $\mathcal{C}$, and samples with noisy labels are considered as the unlabeled set $\mathcal{U}$. For convenience, we have omitted the parameters in the subsequent statements. 

\subsection{Parallel Sample Division}
\begin{figure}[t]
\centering
\includegraphics[scale=0.75]{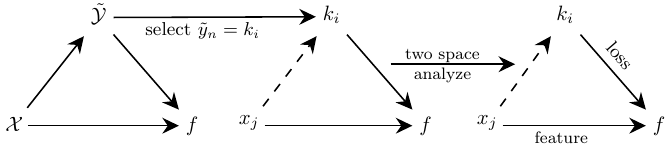}
\caption{Causal diagram of data division in solving the noisy label learning problem. By selecting all samples with the same label as $k_i$ in the label set $\mathcal{\tilde{Y}}$, we are able to analyze the impact of labels $k_i$ and images $x_j$ on the network prediction results $f$ in terms of loss and feature respectively.}
\label{ci}
\end{figure}

Considering the simultaneous exploration in loss space and feature space, we first analyze the rationality of their collaborative utilization from a causal inference perspective. As shown in Fig.~\ref{ci}, both the input image set $\mathcal{X}$ and the noisy label set $\mathcal{\tilde{Y}}$ are factors that influence the output of the network $f$, where the input image set $\mathcal{X}$ is a determining factor for the noisy label set $\mathcal{\tilde{Y}}$. In practice, it is challenging to determine whether the fluctuation in $f$ is due to variations in $\mathcal{X}$ or $\mathcal{\tilde{Y}}$. To address this problem, we select samples $(x_j,k_i)\in\{(x_n,\tilde{y}_n)|\tilde{y}_n = k_i, k_i\in \mathcal{K}, (x_n,\tilde{y}_n)\in\mathcal{D}\}$ to
partition the $K$-class dataset into $K$-noisy cluster as $\mathcal{{D}} = \bigcup_{i=1}^{K} \mathcal{{D}}_i$, where $\mathcal{{D}}_i = \{(x_j, k_i) \}_{j=1}^{|\mathcal{D}_i|}$.
This human intervention effectively fixes $\mathcal{\tilde{Y}}$ to $k_i$, thus establishing a clear and consistent causal relationship between $\mathcal{X}$ and the response variable $\mathcal{\tilde{Y}}$~(an intervention step in causal inference). As a result, we can examine the individual influences of $x_j$ and $k_i$, on the predicted $f$ within each noisy cluster. In particular, we analyze this influence from both loss and feature space, which are explained as follows:
\begin{itemize}
    \item On the one hand, we suppose that the feature extractor is unbiased in the training process, therefore it will have similar responses among samples with clean labels in each cluster, while it will have different responses between one sample with clean label and another sample with noisy label. That is to say, it is possible to distinguish different types of sample by exploring the sample structure in feature space.
    \item On the other hand, we further suppose that the classifier is unbiased in the training process; therefore, the predictions of samples with clean labels will match their labels, while the predictions of samples with noisy labels will mismatch their labels. In other words, it is also possible to distinguish different types of sample by exploring the human prior in loss space.
\end{itemize}
Based on the above analysis, we conduct sample division in the following two steps. First, we model two sample distributions in both feature and loss space to represent the sample structure and human prior, respectively. Second, we distill the training samples into two sets, \emph{i.e.}, \emph{certain} set and \emph{uncertain} set, by conducting the sample clustering with the sample distribution in dual space.

\textbf{Dual-space Sample Distribution.} In loss space, we model the sample distribution by exploring the difference between the predictions of the network and the given noisy labels, whose goal is to further conduct the sample division by finding an optimal criterion from the resulting loss distribution. In particular, we calculate the cross-entropy distribution for each sample $(x_j,k_i)$ within each noisy cluster $\mathcal{D}_i$, which is formulated as follows:
\begin{equation}
    \mathcal{L}_j^p=-{k_{i}}^{T}\log\hat{{y}}_{j},\label{2}
\end{equation}
where ${\hat{y}}_{j}=\mathrm{softmax}(g(f({x}_{j})))$ denotes the predicted probability for each $x_j$ in $(x_j,k_{i})$. In practice, the small-loss criterion is widely used to conduct sample division, in which the sample with a small loss is considered to be the one with a clean label, while the sample with a large loss is regarded as the one with a noisy label.

In feature space, we model the sample distribution by exploring the difference of pairwise similarity for all samples within each noisy cluster, to conduct sample division by finding an optimal criterion from the resulting feature distribution. It is often assumed that the pairwise similarity between intra-class samples is much larger than that between inter-class samples. This assumption inspires us to explore the sample structure within each noisy cluster. In particular, we first compute the category center for each noisy cluster as follows:
\begin{equation}
 \mathcal{O}_c = \frac{1}{N_c}\sum\limits_{j=1}^{N_c} h(f(x_j)), \label{center}
\end{equation}
where $N_c = |\mathcal{D}_i|$ denotes the number of samples with each noisy cluster. Then, we calculate the cosine similarity between each sample and its category center as follows:
\begin{equation}
    \mathcal{L}_j^s =\frac{h(f(x_j)) \cdot \mathcal{O}_{c}}{\Vert h(f(x_j)) \Vert \cdot \Vert  \mathcal{O}_c \Vert}.
\label{sim}
\end{equation}

As a result, we can divide the entire training sample into two different sets, \emph{i.e.}, \emph{certain} set and \emph{uncertain} set, by analyzing the difference of cosine similarity between two samples with clean labels and one sample with clean label and another with noisy label.

\textbf{Dual-space Sample Distillation.} To conduct sample division by analyzing the distributions of the samples in dual space, we apply a clustering algorithm to divide the training samples into \emph{certain} set and \emph{uncertain} set in each noisy cluster. Without loss of generality, we assume that both $ \mathcal{L}_j^p$ and $\mathcal{L}_j^s$ follow a Gaussian mixture distribution, therefore we can take two GMMs~\cite{liu2022gmm} to conduct sample division, which can be formulated as follows:
\begin{equation}
    \mathcal{P}_j^p = \mathtt{GMM}(\mathcal{L}^p_j),
    \hspace{0.2cm}
    \mathcal{P}_j^s = \mathtt{GMM}(\mathcal{L}^s_j),
\end{equation}
where $\mathcal{P}_j^p$ and $\mathcal{P}_j^s$ denote the posterior probabilities of sample $(x_j,k_i)$ belonging to the positive one in loss space and feature space, respectively. In addition, we take two thresholds $t_1$ and $t_2$ to filter out those samples with low confidence, which can be defined as follows:
\begin{equation}
    \mathcal{S}_{p_i}^p = \{ (x_j,k_i) | \mathcal{P}_j^p > t_1 \}_{j=1}^{N_c},
    \mathcal{S}_{p_i}^s = \{ (x_j,k_i) | \mathcal{P}_j^s > t_2 \}_{j=1}^{N_c},
    \label{pos}
\end{equation}
where $\mathcal{S}_{p_i}^p$ and $\mathcal{S}_{p_i}^s$ represent the resulting set of positive samples in loss space and feature space, respectively. As a result, their quality can be significantly maintained in the training process. Similarly, the quality of negative samples can be kept by using the same approach, which can be formulated as follows:
\begin{equation}
    \mathcal{S}_{n_i}^p = \{(x_j,k_i)| \mathcal{P}_j^p \leq t_1 \}_{j=1}^{N_c},
    \mathcal{S}_{n_i}^s = \{(x_j,k_i) | \mathcal{P}_j^s \leq t_2 \}_{j=1}^{N_c},
    \label{neg}
\end{equation}
where $\mathcal{S}_{n_i}^p$ and $\mathcal{S}_{n_i}^s$ represent the resulting set of negative samples in loss space and feature space, respectively. 

To further improve the quality of positive and negative samples, we combine positive samples and negative samples in loss space and feature space, which can be formulated as follows:
\begin{equation}
    \mathcal{S}_p = \bigcup_{i=1}^K \mathcal{S}_{p_i}^p \cap \mathcal{S}_{p_i}^s,
    \hspace{0.2cm}
    \mathcal{S}_n = \bigcup_{i=1}^K \mathcal{S}_{n_i}^p \cap \mathcal{S}_{n_i}^s,
\end{equation}
As a result, the \emph{certain} set and \emph{uncertain} set can be defined as follows:
\begin{equation}
    \mathcal{S}_c = \mathcal{S}_p \cup \mathcal{S}_n,
    \hspace{0.2cm}
    \mathcal{S}_u = \mathcal{D} - \mathcal{S}_c.
\end{equation}
\subsection{Meta Sample Purification}
The quality of samples in the \emph{certain} set can be vigorously guaranteed after parallel sample division, while those samples are less effective in optimizing the parameters of DNN~\cite{NEURIPS2021_ac56f8fe}. As a result, it is necessary to mine more valuable samples in the \emph{uncertain} set to enhance the representation capability of DNN. In practice, hard samples are much more important than easy samples in network training, because the current network has enough ability to handle easy samples but still lacks the ability to handle hard samples. However, because of the limited network representation capability, it is very challenging to directly mine hard samples in the training process. Instead, we plan to gradually mine semi-hard samples via meta sample purification, so as to consistently enhance the network's representation capability across iterations. The meta sample purification module is mainly consisted of two parts, \emph{i.e.}, sample purification modeling and meta-distribution mapping, which are explained in the following paragraphs.

\textbf{Sample Purification Modeling.} The main issue of meta-sample purification lies in how to mine valuable semi-hard samples from $\mathcal{S}_u$. One of the direct choices is to regard the pair of posterior probabilities $\mathcal{P}_n^p$ and $\mathcal{P}_n^s$ as a two-dimensional score $[\mathcal{P}_n^p, \mathcal{P}_n^s]$, and then design a suitable model to learn an appropriate partitioning criterion, which can be simply formulated as follows:
\begin{equation}
    \mathcal{P}_n^f = \mathbf{M}([\mathcal{P}_n^p, \mathcal{P}_n^s]),
    \label{m}
\end{equation}
where $\mathcal{P}_n^f$ denotes a one-dimensional score $\mathcal{P}_n^f$ which can be further used for semi-hard sample mining via an additional threshold filtering. In addition, $\mathbf{M}(\cdot)$ indicates a mapping function that converts the two-dimensional score into a one-dimensional score. In practice, the simplest mapping function is a weighted average of two probabilities, which can be defined as follows:
\begin{equation}
    \mathbf{M}([\mathcal{P}_n^p, \mathcal{P}_n^s]; \lambda) = \lambda \mathcal{P}_n^p + (1 - \lambda) \mathcal{P}_n^s,
    \label{average}
\end{equation}
where $\lambda$ is a constant weight. This form of mapping function only considers a linear relationship between two subsequent probabilities, which is unable to model the nonlinear relationship in some complex suits. Worse still, it is also a very challenging issue of how to choose a suitable weight $\lambda$ in practice, which will in turn decrease the performance in semi-hard sample mining. To address these challenges, we propose a complete meta-distribution mapping solution, which can learn an optimal mapping function to address the semi-hard sample mining problem.
\begin{figure}[t]
\centering
\includegraphics[scale=0.75]{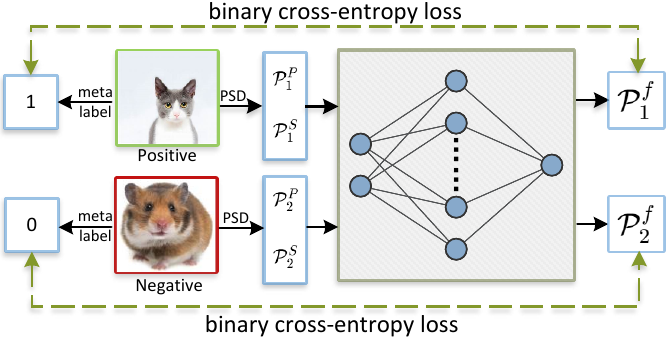}
\caption{Architecture of our meta network, in which:~\textbf{(1)} We take a simple two-layer MLP as the structure of our mapping function;~\textbf{(2)} We take the positive and negative samples in the \emph{certain} set as our meta data in the training process.}
\label{meta}
\end{figure}

\textbf{Meta Distribution Mapping.} The prior meta-learning strategies~\cite{8953770,Zheng_Awadallah_Dumais:2021,Shu_Xie_Yi:2019} often take another set of samples with clean labels as meta-data in the training process. To alleviate this dilemma, we take the positive and negative samples in the \emph{certain} set as our meta-data, and then learn a mapping function to conduct sample purification. On the one hand, the labels of samples in the \emph{certain} set are very accurate, because both the
sample structure in feature space and the human prior in loss space are explored in our parallel sample division. On the other hand, the distribution of our meta-data is the same as that of the training data, because they are directly mined from the original dataset. Once the meta data is ready for model training, we further fine-tune a meta network $\mathbf{m}$ to align with the nonlinear mapping function. According to the Universal Approximation Theorem~\cite{nishijima2021universal}, we adopt a multi-layer perceptron (MLP) in Fig.~\ref{meta} to obtain the optimal nonlinear mapping function. In particular, the network receives the two-dimensional score $[\mathcal{P}_n^p, \mathcal{P}_n^s]$ as input and then maps them to the one-dimensional score $\mathcal{P}_n^f$. In addition, we assign additional binary labels to those meta-samples in the training process. For $(x_n,\tilde{y}_n)$ in the \emph{certain} set, its binary label $b_n$ is given as follows:
\begin{equation}
     b_n=\left\{
\begin{array}{rcl}
1      &  ,    & {(x_n,\tilde{y}_n) \in \mathcal{S}_p},\\
0    &    ,  & {(x_n,\tilde{y}_n) \in \mathcal{S}_n }.\\
\end{array} \right. 
\end{equation}
Afterwards, these meta-data will be used to calculate the binary cross-entropy loss
along with the predicted one-dimensional score $\mathcal{P}_n^f$ as follows:
\begin{equation}
\mathcal{L}_{bce} = b_n\cdot \log \mathcal{P}_n^f+(1-b_n)\cdot \log(1-\mathcal{P}_n^f).
\label{BCE}
\end{equation}
Given the above configuration, the optimization task can be defined as:
\begin{equation}
   \mathbf{m}^{*}(\Theta)=\arg\min_{\mathbf{m}}\sum_{j=1}^{N}\mathcal{L}_{bce}(\mathbf{m}([\mathcal{P}_n^p, \mathcal{P}_n^s];\Theta),b_n),
\end{equation}
where $\Theta$ denotes the optimized parameter.
As a common practice, we utilize the well-known Stochastic Gradient Descent~(SGD) algorithm to minimize loss in the training process. 
 And we can fit the approximate mapping function as $\mathbf{M}\approx\mathbf{m}^{*}(\Theta)$. Furthermore, we can estimate the transformation of a two-dimensional score into a one-dimensional one as follows:
\begin{equation}
    \mathcal{P}_n^f = \mathbf{m}^*([\mathcal{P}_n^p, \mathcal{P}_n^s];\Theta).
\end{equation}
It is obvious that $ \mathcal{P}_n^f \in [0,1]$, in which the higher it is, and the greater the probability that its associated samples have accurate labels. For $(x_n,\tilde{y}_n) \in \mathcal{S}_u$, the final sample purification can be performed by setting two thresholds $t_3$ and $t_4$ to $\mathcal{P}_n^f$, which can be formulated as follows:
\begin{equation}
    \mathcal{C}_u = \{ (x_n,\tilde{y}_n)| \mathcal{P}_n^f \geq t_3 \}_{n=1}^{N_u},
    \hspace{0.2cm}
    \mathcal{U}_u = \{ (x_n,\tilde{y}_n) | \mathcal{P}_n^s \leq t_4 \}_{n=1}^{N_u},
\end{equation}
where $\mathcal{C}_u$ and $\mathcal{U}_u$ denote the samples with clean labels and the samples with noisy labeles, and $N_u = |\mathcal{S}_u|$ denotes the number of samples in the \emph{uncertain} set. Ultimately, the whole dataset is partitioned into two sections: the dataset that contains accurate labels $\mathcal{C}$ and the dataset containing incorrect labels $\mathcal{U}$, with $\mathcal{C} = \mathcal{C}_u \cup \mathcal{S}_p$ and $\mathcal{U} = \mathcal{U}_u \cup \mathcal{S}_n$.

\begin{algorithm}[t]
\caption{Two-Stream Sample Distillation}\label{algorithm}
\KwIn{Training dataset $\mathcal{{D}}$, Meta-Net $\mathbf{m}$, The division thresholds $t_1, t_2, t_3, t_4$}
\KwOut{Almost clean labeled samples $\mathcal{C}$, almost noisy labeled samples  $\mathcal{U}$}
$\mathcal{{D}} \leftarrow \bigcup_{i=1}^{K} \mathcal{{D}}_i$\;
\tcp{PSD}
\For{i=1, ... ,K}{
\For{j=1, ... ,$|\mathcal{{D}}_i|$}{
Obtain $\mathcal{L}_j^p, \mathcal{L}_j^s$ on Eq.~(\ref{2}) and Eq.~(\ref{sim})\;
$\mathcal{P}_j^p \leftarrow \mathtt{GMM}(\mathcal{L}_j^p)$, $\mathcal{P}_j^s \leftarrow \mathtt{GMM}(\mathcal{L}_j^s)$\;
}
Obtain $\mathcal{S}_{p_i}^p, \mathcal{S}_{p_i}^s, \mathcal{S}_{n_i}^p, \mathcal{S}_{n_i}^s$ on Eq.~(\ref{pos}) and Eq.~(\ref{neg})\;
$\mathcal{S}_p \leftarrow \mathcal{S}_p \cup (\mathcal{S}_{p_i}^p \cap \mathcal{S}_{p_i}^s)$,
$\mathcal{S}_n \leftarrow \mathcal{S}_n \cup (\mathcal{S}_{n_i}^p \cap \mathcal{S}_{n_i}^s)$\;
}
$\mathcal{S}_c \leftarrow \mathcal{S}_p \cup \mathcal{S}_n$ and
$\mathcal{S}_u \leftarrow \mathcal{D} - \mathcal{S}_c$\;
\tcp{MSP}
\For{epoch in total epochs}
{\For{$[\mathcal{P}_n^p, \mathcal{P}_n^s]$ of $(x_n,\tilde{y}_n)$ in $\mathcal{S}_c$}
{$\mathcal{P}_n^f \leftarrow \mathbf{m}([\mathcal{P}_n^p, \mathcal{P}_n^s])$\;
\text{Calculate} $\mathcal{L}_{bce}$ on Eq.~(\ref{BCE})\;
$\Theta \leftarrow SGD(\mathcal{L}_{bce},\Theta)$\;
}
}
$\{\mathcal{P}_n^f\}_{n=1}^N\leftarrow \{\mathbf{m}^*([\mathcal{P}_n^p, \mathcal{P}_n^s])\}_{n=1}^N$\;
$\mathcal{C} \leftarrow  \{\mathcal{P}_n^f \geq t_3\}_{n=1}^N$\;
$\mathcal{U} \leftarrow  \{\mathcal{P}_n^f \leq t_4\}_{n=1}^N$\;
\end{algorithm}

We summarize the whole process of our TSSD method in Algorithm~\ref{algorithm}, through which we can acquire an almost clean labeled set $\mathcal{C}$ and a nearly noisy labeled set $\mathcal{U}$. These two sets of training samples are very valuable for the subsequent semi-supervised learning, which is introduced in the next section.
\subsection{Semi-Supervised Learning}
\label{SSL}
Following DivideMix~\cite{li2020dividemix}, we designate the clean labeled set $\mathcal{C}$ as the labeled set $\mathcal{C}$, while omitting the labels from the noisy labeled set $\mathcal{U}$ to serve as the unlabeled set $\mathcal{U}$ to train the network. In the case of a labeled sample $(x_n,\tilde{y}_n)$, we adjust the original label $\tilde{y}_n$ based on probability $\mathcal{P}_{n}^f$ and the average prediction result of the co-teaching networks $p_n$ as follows:
\begin{equation}
    \bar{y}_n = \mathcal{P}_{n}^f \tilde{y}_n + (1 - \mathcal{P}_{n}^f)p_n,
\end{equation}
where $\bar{y}_n$ denotes the refined label. In the case of an unlabeled sample, we use the ensemble of average prediction from co-teaching networks to ``co-guess'' the label $q_n$. After the above operations, we get the augmented dataset $\hat{\mathcal{C}} = \{(x_n,\bar{y}_n)\}_{n=1}^{|\mathcal{C}|}$ and $\hat{\mathcal{U}} = \{(x_n,q_n)\}_{n=1}^{|\mathcal{U}|}$.
Next, we apply the MixMatch method \cite{Berthelot_Carlini_Goodfellow:2019} to convert $\hat{\mathcal{C}}$ and $\hat{\mathcal{U}}$ into $\mathcal{C}'$ and $\mathcal{U}'$.
The loss on $\mathcal{C}'$ is the cross-entropy loss and the loss on $\mathcal{U}'$
is the mean squared error:
\begin{equation}
    \begin{aligned}
&\mathcal{L}_{\mathcal{C}} =-\frac{1}{|{\cal C}^{\prime}|}\sum_{x,p\in{\cal C}^{\prime}}\sum_{k}p_{k}\log(\mathrm{p}_{\mathrm{model}}^{\mathrm{k}}(x;\theta)),  \\
&\mathcal{L}_{\mathcal{U}} =\frac{1}{|\mathcal{U}^{\prime}|}\sum_{x,p\in\mathcal{U}^{\prime}}\|p-\mathrm{p}_{\mathrm{model}}(x;\theta)\|_{2}^{2}. 
\end{aligned}
\end{equation}
To prevent assigning all samples to a single class, we further apply a regularization term~\cite{tanaka2018joint,ICML2019_UnsupervisedLabelNoise} in the training process, which is defined as follows: 
\begin{equation}
\mathcal{L}_{\mathrm{reg}}=\sum_{k}\frac1k\log(\frac1k\frac1{|\mathcal{C}^{\prime}|+|\mathcal{U}^{\prime}|}\sum_{x\in\mathcal{C}^{\prime}+\mathcal{U}^{\prime}}\mathrm{p}_{\mathrm{model}}^{\mathrm{k}}\left(x;\theta\right)).
\end{equation}
Finally, the total loss can be formulated as follows:
\begin{equation}
    {\mathcal{L}}={ \mathcal{L}}_{\mathcal{C}}+\lambda_{u}{ \mathcal{L}}_{\mathcal{U}}+\lambda_{r}{ \mathcal{L}}_{\mathrm{reg}}.
\end{equation}
where $\lambda_{u}$ and $\lambda_{r}$ denote two constant weights.

\section{Experiments}\label{sec:experiments}
\subsection{Datasets}
We evaluate our approach's effectiveness on four benchmark datasets: CIFAR-10/100~\cite{Krizhevsky2009LearningML}, Tiny-ImageNet~\cite{Chrabaszcz2017ADV} and a real-world dataset, Clothing-1M~\cite{Xiao2015LearningFM}, which are introduced as follows:

\textbf{CIFAR-10/100.} The CIFAR-10/100 datasets consist of 50,000 training images and 10,000 test images, respectively. Our experiments examine two types of noise models: symmetric and asymmetric. Specifically, symmetric noise is generated by randomly replacing the labels of a sample portion ($r$) with all possible labels. The design of asymmetric label noise replicates real mistakes, where labels are only substituted with similar classes (e.g., bird $\rightarrow$ airplane, deer $\rightarrow$ horse).

\textbf{Tiny-ImageNet.} Tiny ImageNet is a smaller version
of the full ImageNet ILSVRC. Tiny ImageNet contains 200 training classes. Each class has 500 images. The test set contains 10,000 images. All images are 64x64 colored ones.

\textbf{Clothing-1M.} Clothing-1M contains 1M clothing images in 14 classes. The dataset has noisy labels as a result of its origin from multiple online shopping websites, resulting in numerous mislabeled samples. For training, validating, and testing, the dataset has 50k, 14k, and 10k images, respectively.

\begin{table*}[tbp]
\centering
\caption{Comparison of the Classification Accuracy ($\%$) of various methods in the presence of symmetric and asymmetric noise on the CIFAR-10 and CIFAR-100 datasets.}
\setlength{\tabcolsep}{3.5mm}{
\begin{tabular}{l|cclccc|cclccc}
\toprule
\multicolumn{1}{l|}{\multirow{3}{*}{\textbf{Methods}}}                                                                 & \multicolumn{6}{c|}{\textbf{CIFAR-10}}                                                                                                                                                                  & \multicolumn{6}{c}{\textbf{CIFAR-100}}                                                                                                                                                                  \\ \cline{2-13} 
\multicolumn{1}{c|}{}                                                                                                                    & \multicolumn{3}{c|}{Sym.}                                                                   & \multicolumn{3}{c|}{Asym.}                                                                       & \multicolumn{3}{c|}{Sym.}                                                                   & \multicolumn{3}{c}{Asym.}                                                                        \\ \cline{2-13} 
\multicolumn{1}{c|}{}                                                                                                                    & 20\%                           & 50\%                           & \multicolumn{1}{l|}{80\%} & 10\%                           & 30\%                           & 40\%                           & 20\%                           & 50\%                            & \multicolumn{1}{l|}{80\%} & 10\%                           & 30\%                           & 40\%                           \\ \hline
CE                                                                                                                                       & 86.8                           & 79.4                           & \multicolumn{1}{l|}{62.9} & 88.8                           & 81.7                           & 76.1                           & 62.0                           & 46.7                           & \multicolumn{1}{l|}{19.9} & 68.1                           & 53.3                           & 44.5                           \\
LDMI~\cite{xu2019ldmi}                                                                                             & 88.3                           & 81.2                           & \multicolumn{1}{l|}{43.7} & 91.1                           & 91.2                           & 84.0                           & 58.8                           & 51.8                           & \multicolumn{1}{l|}{27.9} & 68.1                           & 54.1                           & 46.2                           \\
MixUp~\cite{zhang2018mixup}                                                                                        & 95.6                           & 87.1                           & \multicolumn{1}{l|}{71.6} & 93.3                           & 83.3                           & 77.7                           & 67.8                           & 57.3                           & \multicolumn{1}{l|}{30.8} & 72.4                           & 57.6                           & 48.1                           \\
Co-teaching+~\cite{yu2019does}                                                                                     & 89.5                           & 85.7                           & \multicolumn{1}{l|}{67.4} & 93.8                           & 92.5                           & 91.7                           & 65.6                           & 51.8                           & \multicolumn{1}{l|}{27.9} & 71.6                           & 69.5                           & 55.1                           \\
DivideMix~\cite{li2020dividemix}                                                                                   & 96.1                           & 94.6                           & \multicolumn{1}{l|}{92.9} & 94.2                           & 94.1                           & 93.2                           & 77.3                           & 74.6                           & \multicolumn{1}{l|}{60.2} & 77.4                           & 75.1                           & 74.0                           \\
U{\footnotesize{NI}}C{\footnotesize{ON}}~\cite{Karim_Rizve_Rahnavard:2022} & 96.0                           & 95.6                           & \multicolumn{1}{l|}{93.9} & 95.3                           & 94.8                           & 94.1                           & 78.9                           & 77.6                           & \multicolumn{1}{l|}{63.9} & 78.2                           & 75.6                           & 74.8                           \\
TCL~\cite{huang2023twin}                                                                                                 & 95.0                           & 93.9                           & \multicolumn{1}{l|}{92.5} & -                              & -                              & 92.6                           & 78.0                           & 73.3                           & \multicolumn{1}{l|}{\textbf{65.0}} & -                              & -                              & -                              \\ \hline
TSSD                                                                                                                                     & \textbf{96.7} & \textbf{95.7} & \multicolumn{1}{l|}{\textbf{95.0}} & \textbf{96.5} & \textbf{96.2} & \textbf{95.1} & \textbf{82.1} & \textbf{78.1} & \multicolumn{1}{l|}{64.2} & \textbf{82.3} & \textbf{78.9} & \textbf{75.4} \\ \bottomrule
\end{tabular}}

\label{table1}
\end{table*}
\begin{table}[ht]
\centering
\caption{Comparison of the Classification Accuracy ($\%$) of various methods in the presence of symmetric noise on the Tiny-ImageNet dataset.}
\label{table3}
\setlength{\tabcolsep}{6mm}
\begin{tabular}{llll}
\toprule
\multicolumn{4}{c}{\textbf{Tiny-ImageNet}}                                                                            \\ \hline
\multicolumn{1}{l|}{Methods}           & 0\%  & 20\% & 50\%                           \\ \hline
\multicolumn{1}{l|}{CE}                                       & 57.4 & 35.8 & 19.8                           \\
\multicolumn{1}{l|}{Decoupling~\cite{NIPS2017_58d4d1e7}}     & - & 37.0 & 22.8 \\                                 
\multicolumn{1}{l|}{F-correction~\cite{patrini2017making}}      & -    & 44.5 & 33.1                           \\
\multicolumn{1}{l|}{MentorNet~\cite{jiang2018mentornet}}        & -    & 45.7 & 35.8                           \\
\multicolumn{1}{l|}{Co-teaching+~\cite{yu2019does}}             & 52.4 & 48.2 & 41.8                           \\
\multicolumn{1}{l|}{M-correction~\cite{ICML2019_UnsupervisedLabelNoise}}  & 57.7 & 57.2 & 51.6                           \\
\multicolumn{1}{l|}{NCT~\cite{sarfraz2020noisy}}                & 62.4 & 58.0 & 47.8                           \\
\multicolumn{1}{l|}{UNICON~\cite{Karim_Rizve_Rahnavard:2022}} & 62.7 & 59.2 & 52.7                           \\ \hline
\multicolumn{1}{l|}{TSSD}                                     & \textbf{63.1} &\textbf{60.9} & \textbf{53.5}\\
\bottomrule
\end{tabular}

\end{table}

\subsection{Training Details}
We employed different training approaches for each dataset. The PreAct ResNet18 architecture ~\cite{He2016IdentityMI} is used for CIFAR-10, CIFAR-100, and Tiny-ImageNet. The ResNet50~\cite{He_Zhang_Ren:2016} network pre-trained on ImageNet is chosen as the backbone for Clothing-1M. Meta-Net and the final classification network are trained using SGD optimization. For all datasets, the Meta-Net learning rate is set to 0.2. Training for CIFAR-10 is conducted over 350 epochs, with a 10-epoch warm-up period. For CIFAR-100, the training spans 350 epochs with a warm-up period of 30 epochs. In both cases, the initial learning rate is set to 0.04 and gradually reduced by 0.1 every 120 epochs. Regarding the Tiny-ImageNet dataset, the training is performed for 350 epochs, with a warm-up phase of 15 epochs. The initial learning rate is set to 0.005 and decays linearly every 120 epochs. For Clothing-1M, the network is trained for 80 epochs, with a 1-epoch warm-up period. The initial learning rate is set at 0.002, and after 40 epochs, the learning rate is reduced by a factor of 10. To augment the data, we employ the AutoAugment Policy~\cite{Cubuk_2019_CVPR}, using the CIFAR-10 Policy for CIFAR-10 and CIFAR-100, and the ImageNet Policy for Tiny-ImageNet and Clothing-1M. The experimental parameters above for training the models were established on the basis of previous work~\cite{Karim_Rizve_Rahnavard:2022,zhao2022centrality}.

The hyperparameters for semi-supervised learning in CIFAR-10 and CIFAR-100 are set as follows: $\mathcal{\lambda}_u$ and $\mathcal{\lambda}_r$ are set to 30 and 1, respectively. Similarly, for Tiny-ImageNet, $\mathcal{\lambda}_u$ is set to 50 and $\mathcal{\lambda}_r$ to 1. For Clothing-1M, $\mathcal{\lambda}_u$ is set to 0, and $\mathcal{\lambda}_r$ to 1. These parameters for semi-supervised learning were inherited from the settings in the U{\footnotesize{NI}}C{\footnotesize{ON}}~\cite{Karim_Rizve_Rahnavard:2022}.
The division thresholds, denoted $t_1, t_2, t_3, t_4$, will be examined and discussed in the ablation studies. All experiments were carried out on NVIDIA GeForce RTX 3090 GPUs.

\begin{table}[tbp]
  \begin{center}
    \caption{Comparison of the Accuracy~($\%$) in Classification of various methods on the Clothing-1M dataset.}
    \setlength{\tabcolsep}{6mm}
    \begin{tabular}{l|c|c} 
    \toprule
    \multicolumn{3}{c}{\textbf{Clothing-1M}}                                                               \\ \hline
      Methods & Backbone & Accuracy\\
      \hline
      CE & ResNet-50 &69.2\\
      Joint-Optim~\cite{tanaka2018joint}  & ResNet-50 &72.0\\
      MetaCleaner~\cite{8953770} & ResNet-50 &72.5\\
      PCIL~\cite{yi2019probabilistic} & ResNet-50 &73.5\\
      DivideMix~\cite{li2020dividemix} & ResNet-50 &74.8\\
      ELR~\cite{liu2020earlylearning}& ResNet-50 &74.8\\
    U{\footnotesize{NI}}C{\footnotesize{ON}}~\cite{Karim_Rizve_Rahnavard:2022}& ResNet-50 &74.9\\
    CC~\cite{zhao2022centrality}& ResNet-50 & 75.4\\
    TCL~\cite{huang2023twin}& ResNet-50 & 74.8\\
    OT-Filter~\cite{10203559}& ResNet-50 & 74.5\\
      \hline
    TSSD &ResNet-50 &\textbf{75.6}\\
    \bottomrule
    \end{tabular}
    
    \label{table4}
  \end{center}
  
\end{table}
\subsection{Comparison with the State-of-the-Art Methods}

This section presents comparative analyses of the classification performance between TSSD and other methods.
 
\textbf{CIFAR-10/100: }Table~\ref{table1} presents the average performance on the CIFAR-10 and CIFAR-100 datasets, considering the symmetric noise levels of $20\%$, $50\%$, and $80\%$, as well as asymmetric noise levels of $10\%$, $30\%$, and $40\%$. Our method exceeds most state-of-the-art (SOTA) methods, particularly exhibiting significant improvements under common noise conditions. However, our results are slightly underperforming compared to the results achieved by TCL~\cite{huang2023twin} in the CIFAR-100 dataset with 80\% symmetric noise. This difference could potentially be attributed to the insufficient accuracy in meta-samples selected from feature space and loss space under higher noise levels, consequently impacting the performance of the meta classifier. As a prospective solution, the incorporation of a small set of clean data could be considered to mitigate this issue.

\textbf{Tiny-ImageNet: }Table~\ref{table3} presents the average performance of the Tiny-ImageNet dataset at symmetric noise levels of $0\%$, $20\%$, and $50\%$. Our method exceeds most SOTA methods. We follow the same training methodology as U{\footnotesize{NI}}C{\footnotesize{ON}}~\cite{Karim_Rizve_Rahnavard:2022}. In contrast to U{\footnotesize{NI}}C{\footnotesize{ON}}, which solely utilizes JS-divergence for sample selection in loss space, our approach incorporates sample selection in feature space as well. This enables us to identify more potential clean samples and improve the accuracy of clean samples using the PSD module. As a result, we observed an approximately 1\% improvement in performance.

\textbf{Clothing-1M: }Table~\ref{table4} presents the average performance of the Clothing-1M dataset in real-world scenarios, demonstrating that our method outperforms SOTA methods and achieves superior results. We use the same training methodology as CC~\cite{zhao2022centrality} but introduce a loss space component, in contrast to CC's use of only feature space. Furthermore, we integrate two space using our proposed MSP module, resulting in a performance improvement of approximately 0.2\% compared to CC. These findings indicate that loss space contains challenging samples not identified solely by feature space filtering approach. The inclusion of these challenging samples significantly contributes to improved performance.

\begin{table*}[htbp]
\centering
\caption{
Comparison between our TSSD with $\text{Baseline}_{\text{loss1}}$ corresponds to the utilization of the cross-entropy method as described in Eq.~\ref{2}, $\text{Baseline}_{\text{loss2}}$ applies JS-divergence and $\text{Baseline}_{\text{feat}}$ corresponds to the feature similarity approach outlined in Eq.~\ref{sim}.
}
\setlength{\tabcolsep}{2.5mm}
\begin{tabular}{l|ccc|ccc|c|cc}
\toprule
\multirow{2}{*}{\textbf{Methods}} & \multicolumn{3}{c|}{\textbf{CIFAR-10}}                                                  & \multicolumn{3}{c|}{\textbf{CIFAR-100}}                                                 & \textbf{Clothing-1M} & \multicolumn{2}{c}{\textbf{Tiny-ImageNet}}                         \\ \cline{2-10} 
                                  & \multicolumn{1}{c|}{20\%-sym.}     & \multicolumn{1}{c|}{50\%-sym.}     & 80\%-sym.     & \multicolumn{1}{c|}{20\%-sym.}     & \multicolumn{1}{c|}{50\%-sym.}     & 80\%-sym.     & -                    & \multicolumn{1}{l|}{20\%-sym.}     & \multicolumn{1}{l}{50\%-sym.} \\ \hline
$\text{Baseline}_{\text{loss1}}$                     & \multicolumn{1}{c|}{96.1}          & \multicolumn{1}{c|}{94.6}          & 92.9          & \multicolumn{1}{c|}{77.3}          & \multicolumn{1}{c|}{74.6}          & 60.2          & 74.8                 & \multicolumn{1}{c|}{58.9}          & 53.1                          \\
$\text{Baseline}_{\text{loss2}}$                     & \multicolumn{1}{c|}{96.0}          & \multicolumn{1}{c|}{94.2}          & 91.7          & \multicolumn{1}{c|}{77.5}          & \multicolumn{1}{c|}{75.7}          & 60.1          & 74.9                 & \multicolumn{1}{c|}{59.2}          & 52.7                          \\ \hline
$\text{Baseline}_{\text{feat}}$                   & \multicolumn{1}{c|}{95.8}          & \multicolumn{1}{c|}{95.5}          & 93.6          & \multicolumn{1}{c|}{80.6}          & \multicolumn{1}{c|}{77.4}          & 60.7          & 74.5                 & \multicolumn{1}{c|}{59.5}          & 52.9                          \\ \hline
TSSD                              & \multicolumn{1}{c|}{\textbf{96.7}} & \multicolumn{1}{c|}{\textbf{95.7}} & \textbf{95.0} & \multicolumn{1}{c|}{\textbf{82.1}} & \multicolumn{1}{c|}{\textbf{78.1}} & \textbf{64.4} & \textbf{75.6}        & \multicolumn{1}{c|}{\textbf{60.9}} & \textbf{53.5}                 \\ \bottomrule
\end{tabular}
\label{ablation}
\end{table*}

\begin{table*}[t]
\centering
\caption{The F1-score of the samples filtered out by the loss space and feature space respectively under different $t_1,t_2$ settings in the CIFAR-100 dataset with varying levels of symmetric noise.}
\setlength{\tabcolsep}{6mm}
\begin{tabular}{l|ccc|cc|ccc}
\toprule
{\textbf{CIFAR-100}} & \multicolumn{3}{c|}{20\%-sym.}                                                & \multicolumn{2}{c|}{50\%-sym.}                   & \multicolumn{3}{c}{80\%-sym.}                                                 \\ \hline
$t_1\ \&\ t_2$         & \multicolumn{1}{c|}{0.2}   & \multicolumn{1}{c|}{0.5}   & 0.2-th         & \multicolumn{1}{c|}{0.5}   & 0.5-th         & \multicolumn{1}{c|}{0.5}   & \multicolumn{1}{c|}{0.8}   & 0.8-th         \\ \hline
$ {\text{F1-score}\ \text{(loss)}}$      & \multicolumn{1}{c|}{0.915} & \multicolumn{1}{c|}{0.894} & \textbf{0.926} & \multicolumn{1}{c|}{0.871} & \textbf{0.875} & \multicolumn{1}{c|}{0.775} & \multicolumn{1}{c|}{0.822} & \textbf{0.830} \\
$ {\text{F1-score}\ \text{(feat)}}$   & \multicolumn{1}{c|}{0.918} & \multicolumn{1}{c|}{0.893} & \textbf{0.931} & \multicolumn{1}{c|}{0.869} & \textbf{0.871} & \multicolumn{1}{c|}{0.784} & \multicolumn{1}{c|}{0.826} & \textbf{0.833} \\ \bottomrule
\end{tabular}
\label{Parameter1}
\end{table*}

\begin{table*}[t]
\centering
\caption{The F1-score of the samples filtered out by combining the loss space and feature space under different $t_3,t_4$ settings in the CIFAR-100 dataset with varying levels of symmetric noise.}
\setlength{\tabcolsep}{6mm}
\begin{tabular}{l|ccc|cc|ccc}
\toprule
{\textbf{CIFAR-100}} & \multicolumn{3}{c|}{20\%-sym.}                                            & \multicolumn{2}{c|}{50\%-sym.}                       & \multicolumn{3}{c}{80\%-sym.}                                             \\ \hline
$t_3\ \&\ t_4$        & \multicolumn{1}{c|}{0.2}            & \multicolumn{1}{c|}{0.5}   & 0.2-th & \multicolumn{1}{c|}{0.5}            & 0.5-th         & \multicolumn{1}{c|}{0.5}   & \multicolumn{1}{c|}{0.8}            & 0.8-th \\ \hline
F1-score  & \multicolumn{1}{c|}{\textbf{0.937}} & \multicolumn{1}{c|}{0.935} & 0.936  & \multicolumn{1}{c|}{\textbf{0.883}} & \textbf{0.883} & \multicolumn{1}{c|}{0.841} & \multicolumn{1}{c|}{\textbf{0.846}} & 0.834  \\  \bottomrule
\end{tabular}
\label{Parameter2}
\end{table*}


\section{Ablation studies}\label{sec:ablation}
In this section, we conduct ablation studies of TSSD in different training settings.

\textbf{Effect of Combining Two Spaces: }In this study, we analyze the effect of combining two spaces on the accuracy of the final test set. Specifically, we individually compare the test accuracy achieved by employing the loss-based methods~(Cross-Entropy and JS-Divergence) and feature-based method~($\mathcal{L}_n^s$) with the test accuracy obtained by our TSSD method. The comparative results are presented in Table~\ref{ablation}. Our TSSD method has improved significantly compared to methods based solely on cross-entropy, JS-divergence, or $\mathcal{L}_n^s$. This indicates that the two-space method can perform better than the filtering method with a single space. This improvement arises because when detection is carried out in two spaces, false positive~(FP) and false negative~(FN) instances created within one space are not classified as noisy labeled instances and correctly labeled instances. Instead, a secondary evaluation is conducted to determine their characteristics. The secondary evaluation process reveals semi-hard samples among false positives and false negatives.
These semi-hard samples offer the network a wealth of valuable information, thus bolstering the network's robustness.

\begin{figure}[t]
\centering
\includegraphics[scale=0.26]{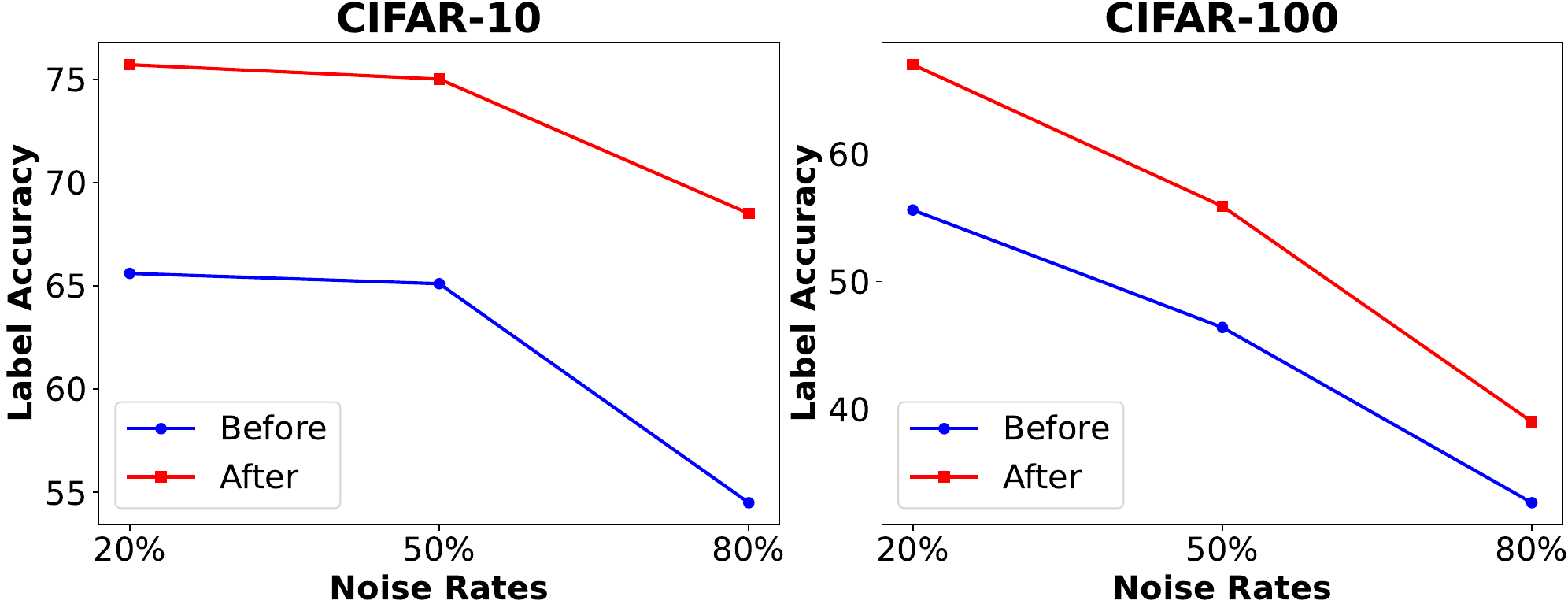}
\caption{Evaluation of label purification. The label accuracy of uncertain samples is compared before and after utilizing MSP on the CIFAR-10/100 datasets. These datasets consist of varying rates of symmetric noise. }
\label{ba}
\end{figure}
\begin{figure*}[t]
\centering
\includegraphics[scale=0.66]{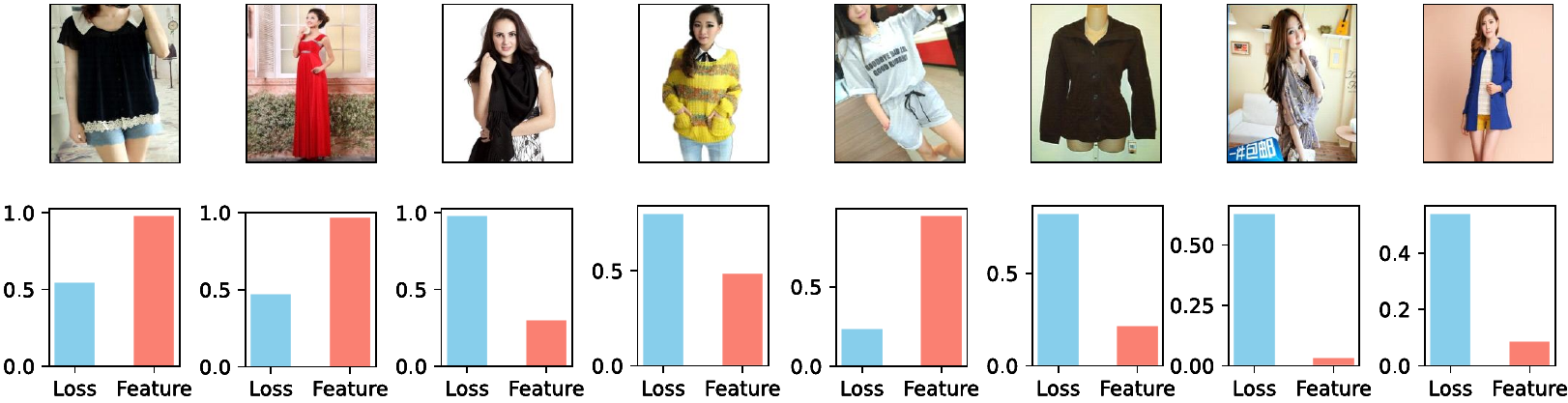}
\caption{Comparison of the potential probability of clean labels for some samples in the Clothing-1M dataset that show discrepancies between loss space and feature space. The sample images on the left correspond to the bar graph on the right based on their respective positions.}
\label{example}
\end{figure*}
\begin{figure}[h]
\centering
\includegraphics[scale=0.23]{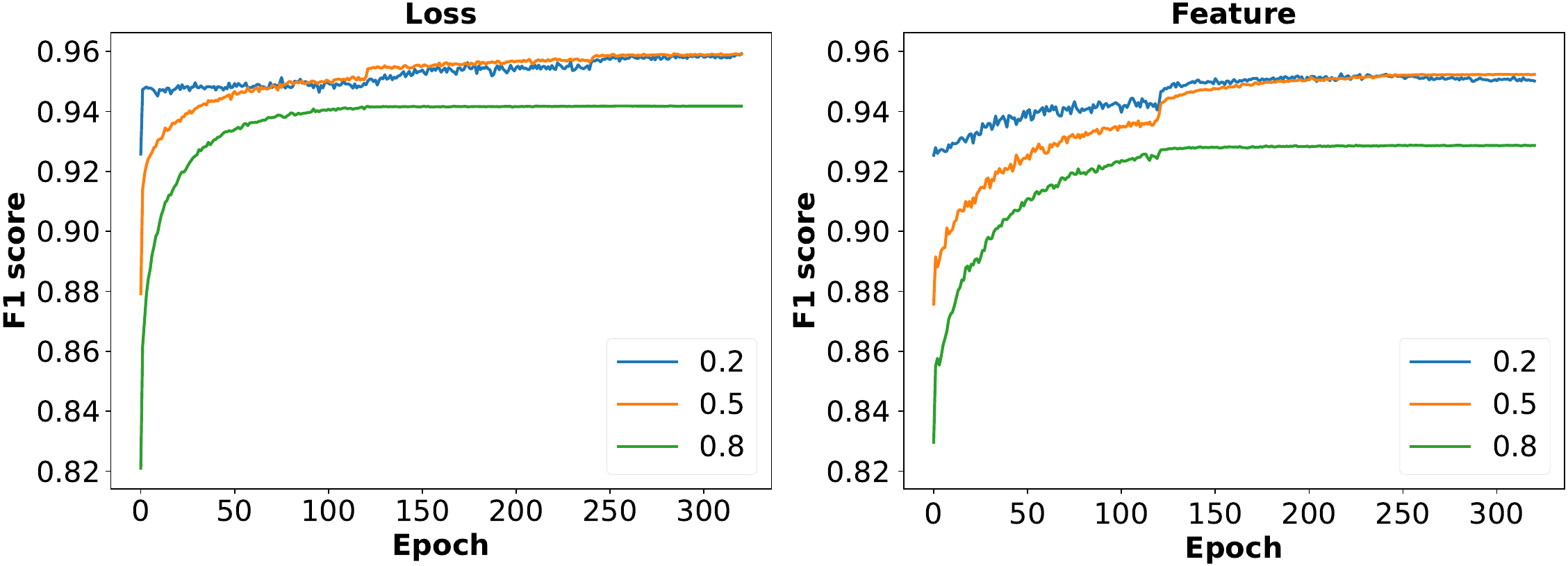}
\caption{The F1-scores of the selected samples from loss space and feature space vary as the training progresses on the CIFAR-100 dataset, which contains 20\%, 50\%, and 80\% levels of symmetric noise.}
\label{f1}
\end{figure}

\textbf{Evaluation of Label Purification: }In this study, we analyze the impact of the MSP module on the \emph{uncertain} set selected by the PSD module. As shown in Fig.~\ref{ba}, we present the precision of clean labeled samples in the original \emph{uncertain} set and the precision of clean labeled samples filtered out of the \emph{uncertain} set using the MSP module. This analytical experiment showcases the results of the first epoch after the completion of warm-up on the CIFAR-10/100 datasets, which consist of varying degrees of symmetric noise. It can be seen that our module improves the cleanliness of \emph{uncertain} set by approximately 10\% at different levels of noise. By significantly improving the cleanliness of the \emph{uncertain} set, we can achieve optimal performance in the final classification.

\textbf{Parameter Analysis: }We explore the impacts of the sample selection threshold $t_1, t_2, t_3, t_4$. We set $t_1 = t_2,\  t_3 = t_4$ to reduce the difficulty of finding parameters. There are three commonly used methods for setting a more reasonable threshold: One is simply to set the threshold to 0.5. Another method is to set the threshold on the basis of the estimated noise level in the dataset. The third method is to estimate the noise rate, denoted $p$, and set the threshold as the $p$-th percentile of the total data. We evaluated the F1-score of selected samples from the CIFAR-100 dataset with varying degrees of label noise. The results for $t_1 = t_2$ are shown in Table~\ref{Parameter1}, and the results for $t_3 = t_4$ are shown in Table~\ref{Parameter2}. As we can see, for $t_1,t_2$, using the p-th percentile of the data allows for a higher quality filtering of the data. However, for datasets with unknown noise levels, using a threshold of 0.5 is also feasible, as it only leads to a decrease in data quality of less than 0.1. For $t_3,t_4$, using the estimated noise level of the data yields better results. However, even for datasets with unknown noise levels, using 0.5 as a threshold does not result in a significant decrease in data quality. In our experimental setup, for the CIFAR-10/100 and Tiny-ImageNet datasets, we use the p-th percentile for $t_1, t_2$ and the noise level for $t_3,t_4$. For the Clothing1m dataset, we set both $t_1, t_2, t_3, t_4$ at 0.5.

\textbf{Verification of Motivation: } \textbf{(1) Difference: }The Difference between dual-space lies primarily in the quality of the data chosen. This quality is mainly determined by the recall rate and the precision rate of the selected samples, with the F1-score taking both into account. Therefore, Fig.~\ref{f1} illustrates the change in the F1-scores of the samples selected from dual-space during training on the CIFAR-100 dataset with varying levels of noise. Although both methods exhibit a similar trend of initial improvement followed by stabilization, the F1-score of samples chosen through the loss method displays faster improvement and achieves a higher score compared to those selected via the feature method. 
\textbf{(2) Complementarity: }In the Clothing-1M dataset, we use the same pre-trained model and use GMM to divide samples according to $\mathcal{L}_n^p$ and $\mathcal{L}_n^s$, respectively. For some sample examples, $\mathcal{P}_n^p$ and $\mathcal{P}_n^s$ in both spaces are shown in Fig.~\ref{example}. Probability presents a high value in one space and a low value in another.  This enables the same sample to produce different division results in each space, and correct segmentation results can complement incorrect segmentation results so that they are not immediately segmented by a single space. The presence of this complementarity is not exclusive, as similar occurrences have been observed in other dataset. As shown in Fig.~\ref{label_mot}, such complementary samples are consistently found. This indicates that complementarity is not inherent in the data, but rather arises from the discrepancy between loss space and feature space.


    
    

\begin{figure*}[t]
\centering
\includegraphics[scale=0.35]{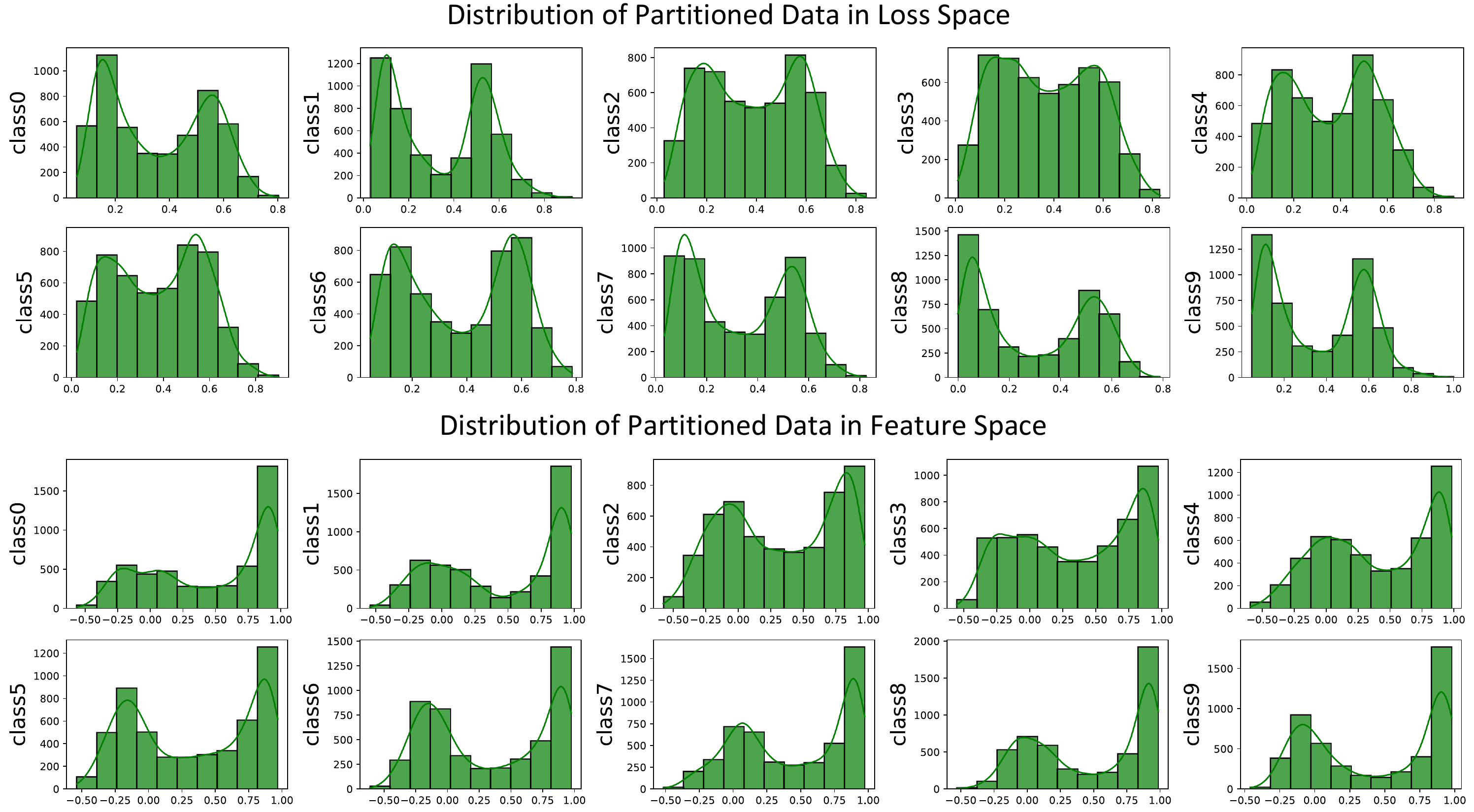}
\caption{The distribution of $\mathcal{L}_n^p$ in loss space and $\mathcal{L}_n^s$ in feature space for each category in the CIFAR-10 dataset with 50\% symmetric noise is presented. The bars in both graphs indicate the frequency of $\mathcal{L}_n^p$ or $\mathcal{L}_n^s$ within each interval. Gaussian Kernel Density Estimation is used to model the overall data distribution. The fitted curves adhere to the Gaussian mixture model. The top row of the subplot pair, moving from left to right, represents classes 1-5 in the dataset, whereas the bottom row, moving from left to right, represents classes 6-10 in the dataset.}
\label{gmm}
\end{figure*}

\textbf{Validating the Use of GMM: }The use of GMM for sample partitioning in both loss-based and feature-based methods has been extensively studied~\cite{li2020dividemix,zhao2022centrality}. However, the main methods mostly directly partition the computed loss or similarity between all samples, while our method focuses on partitioning each class separately. Therefore, we conducted an investigation of the distribution of $\mathcal{L}_n^p$ and $\mathcal{L}_n^s$ for each class in the CIFAR-10 dataset with 50\% symmetric noise. Fig.~\ref{gmm} shows the data distribution of $\mathcal{L}_n^p$ for each class in loss space and the data distribution of $\mathcal{L}_n^s$ for each class in feature space. The bars in the two plots represent the count of $\mathcal{L}_n^p$ or $\mathcal{L}_n^s$ in each range. The overall data distribution is modeled using Gaussian Kernel Density Estimation. The curves that have been fitted adhere to the Gaussian mixture model and exhibit a distinct bimodal nature, offering a foundation for partition the data using the GMM.

\begin{table}[t]
\centering
\caption{Comparison of the quality of data and the accuracy of classification achieved through the MSP method and the weighted average method on the CIFAR-100 dataset at varying levels of noise rates.}
\setlength{\tabcolsep}{1.5mm}
\begin{tabular}{ll|ccc|c}
\toprule
\multicolumn{2}{l|}{\multirow{2}{*}{\textbf{CIFAR-100}}}              & \multicolumn{3}{c|}{\textbf{Average+PSD}}                                               & \multirow{2}{*}{\textbf{MSP+PSD}} \\ \cline{3-5}
\multicolumn{2}{l|}{}                           & \multicolumn{1}{l}{$\lambda = 1$} & \multicolumn{1}{l}{$\lambda = 0$} & $\lambda = 0.5$ &                                   \\ \hline
\multicolumn{1}{c|}{\multirow{3}{*}{20\%-Sym.}} & TP rate             & 96.2                              & 95.7                              & 96.3            & \textbf{96.6}                     \\
\multicolumn{1}{c|}{}                           & TN rate             & 85.1                              & 81.9                              & 84.9            & \textbf{85.6}                     \\ \cline{2-6} 
\multicolumn{1}{c|}{}                           & classification acc. & 77.3                              & 80.6                              & 79.5            & \textbf{82.1}                     \\ \hline
\multicolumn{1}{l|}{\multirow{3}{*}{50\%-Sym.}} & TP rate             & 88.0                              & 87.6                              & 88.7            & \textbf{88.9}                     \\
\multicolumn{1}{l|}{}                           & TN rate             & 87.0                              & 86.7                              & 87.7            & \textbf{87.8}                     \\ \cline{2-6} 
\multicolumn{1}{l|}{}                           & classification acc. & 74.6                              & 77.4                              & 77.6            & \textbf{78.1}                     \\ 
\bottomrule
\end{tabular}
\label{selection}
\end{table}

\textbf{MSP vs. Weighted Average: }We compare the MSP method with the weighted average method using various $\lambda$ values (0, 1, and 0.5) in Eq.~(\ref{average}). Evaluation involves evaluating the precision of classification and the true positive (TP) and true negative (TN) rates in data partitioning, as described in Table~\ref{selection}. The enhanced classification may arise from the fact that the MSP method eliminates more precise data points. Although the enhancement in TP and TN rates within a single epoch may be marginal, this enhancement progressively accumulates over successive training epochs, leading to an overall improvement in the network's performance. Consequently, a more effective network results in better classification accuracy.

\section{Conclusions}\label{sec:conclusion}
We propose a Two-Stream Sample Distillation framework comprising two modules to tackle the problem of learning from noisy labels. The first Parallel Sample Division module generates high-fidelity positive and negative sets by jointly considering the sample structure in feature and loss space. The second Meta Sample Purification module further judges samples in the \emph{uncertain} set by learning a solid meta-classifier on positive and negative sets. We conducted extensive experiments on several challenging datasets to demonstrate the effectiveness of our method in better exploring semi-hard samples and providing more accurate sample purification. 

\textbf{Limitations and Future Work}. A limitation of our approach is its exclusive focus on feature space and loss space. However, it is essential to note that these two metrics are only one of many evaluation criteria for data selection. In future investigations, it would be advantageous to incorporate information from multiple metrics to improve the quality of selection results.






\bibliographystyle{IEEEtran}
\bibliography{TSSD}

\ifCLASSOPTIONcaptionsoff
  \newpage
\fi

\begin{IEEEbiography}
[{\includegraphics[width=1in,height=1.25in,clip,keepaspectratio]{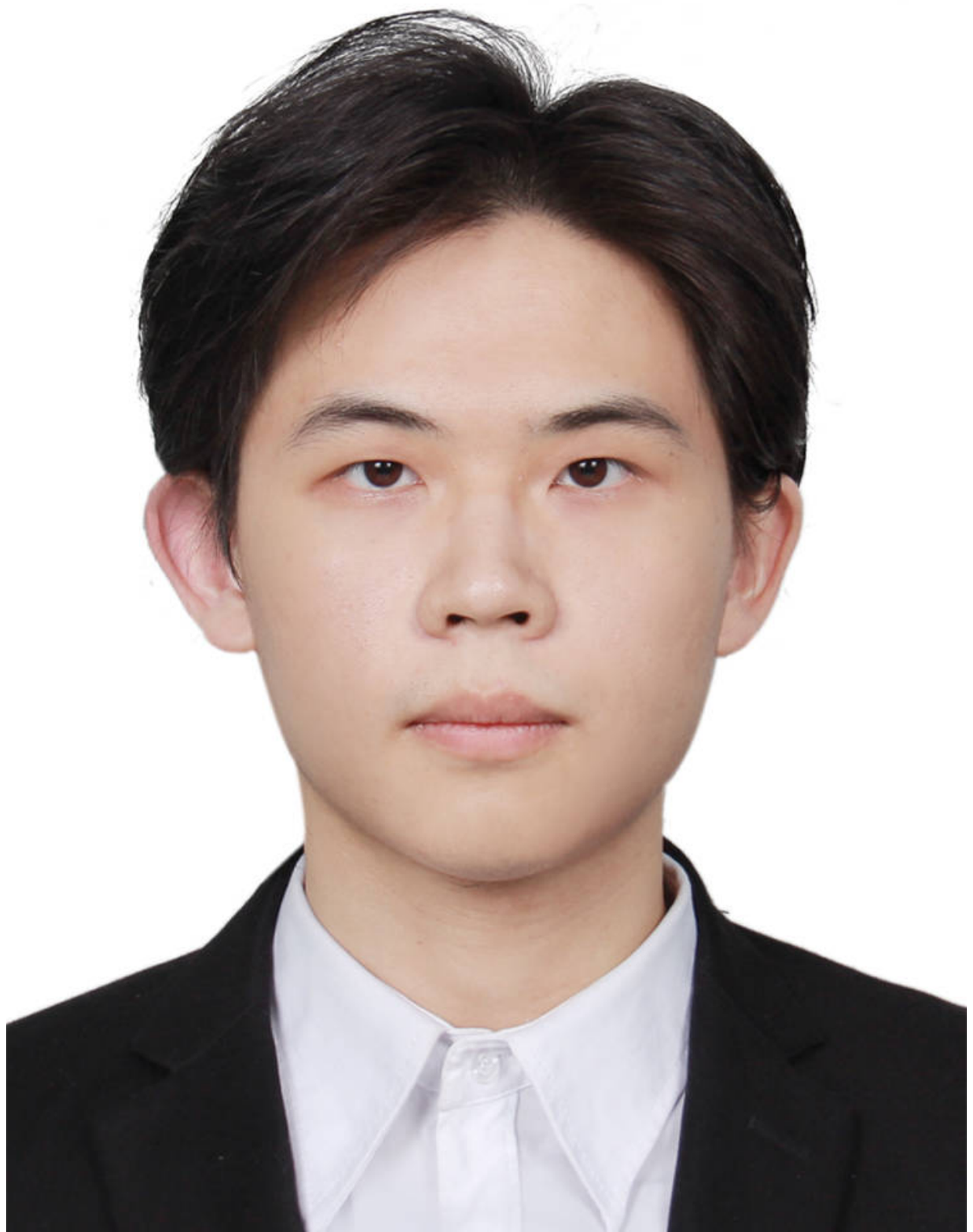}}]{Sihan Bai}
received the B.E. degree in the School of Information and Computational Science, Harbin Institute of Technology, China, in 2022.
He is currently pursuing the M.E. degree at the Institute of Artificial Intelligence and Robotics of Xi'an Jiaotong University. His research interests include computer vision and machine learning.
\end{IEEEbiography}

\begin{IEEEbiography}[{\includegraphics[width=1in,height=1.25in,clip,keepaspectratio]{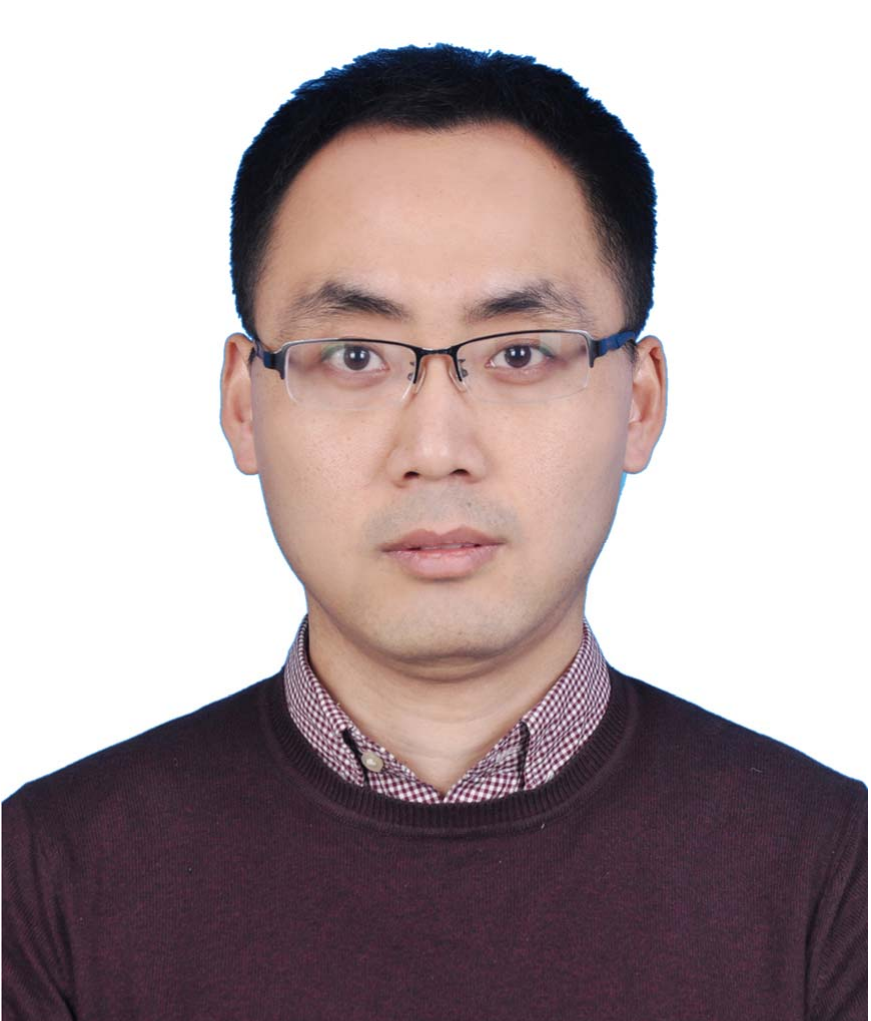}}]{Sanping Zhou} (Member, IEEE)
received the Ph.D. degree in control science and engineering from Xi'an Jiaotong University, Xi'an, China, in 2020. From 2018 to 2019, he was a visiting Ph.D. student with the Robotics Institute, Carnegie Mellon University. He is currently an Associate Professor with the Institute of Artificial Intelligence and Robotics, Xi'an Jiaotong University. His research interests include machine learning, deep learning, and computer vision, with a focus on person re-identification, salient object detection, medical image segmentation, image classification, and visual tracking.
\end{IEEEbiography}

\begin{IEEEbiography}
[{\includegraphics[width=1in,height=1.25in,clip,keepaspectratio]{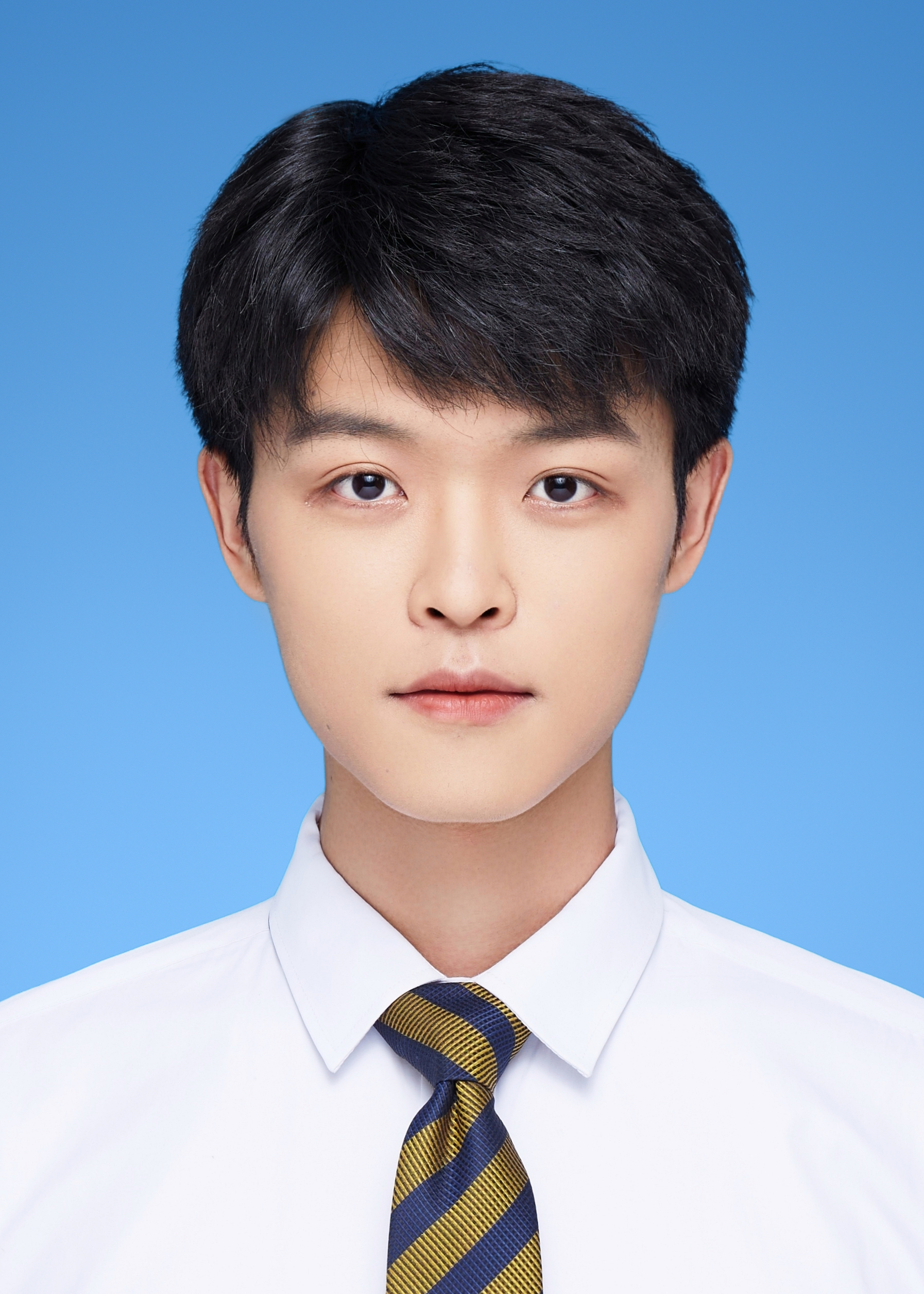}}]{Zheng Qin}
received the B.S. degree in robotic engineering from the Harbin Institute of Technology, China, in 2021. He is currently working toward the PdD. degree in artificial intelligence from Xi'an Jiaotong University. His research interests include multi-object tracking, embodied intelligence and visual navigation.
\end{IEEEbiography}

\begin{IEEEbiography}[{\includegraphics[width=1in,height=1.25in,clip,keepaspectratio]{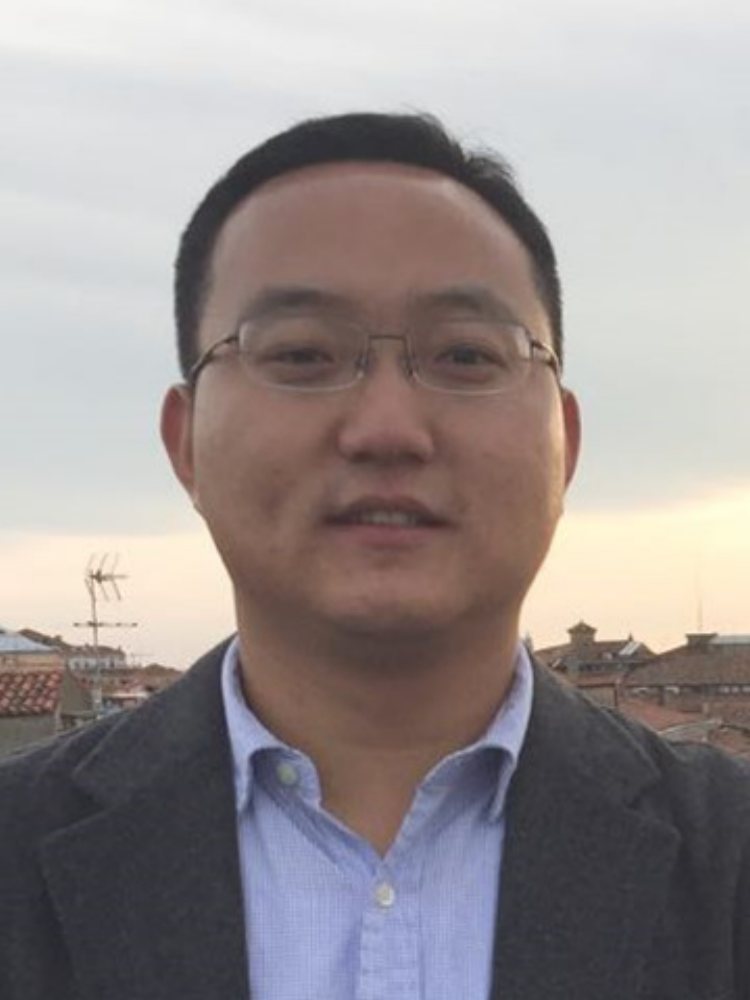}}]
{Le Wang}(Senior Member, IEEE) received the B.S. and Ph.D. degrees in Control Science and Engineering from Xi'an Jiaotong University, Xi'an, China, in 2008 and 2014, respectively. From 2013 to 2014, he was a visiting Ph.D. student with Stevens Institute of Technology, Hoboken, New Jersey, USA. From 2016 to 2017, he was a visiting scholar with Northwestern University, Evanston, Illinois, USA. He is currently a Professor with the Institute of Artificial Intelligence and Robotics of Xi'an Jiaotong University, Xi'an, China. His research interests include computer vision, pattern recognition, and machine learning.
\end{IEEEbiography}


\begin{IEEEbiography}[{\includegraphics[width=1in,height=1.25in,clip,keepaspectratio]{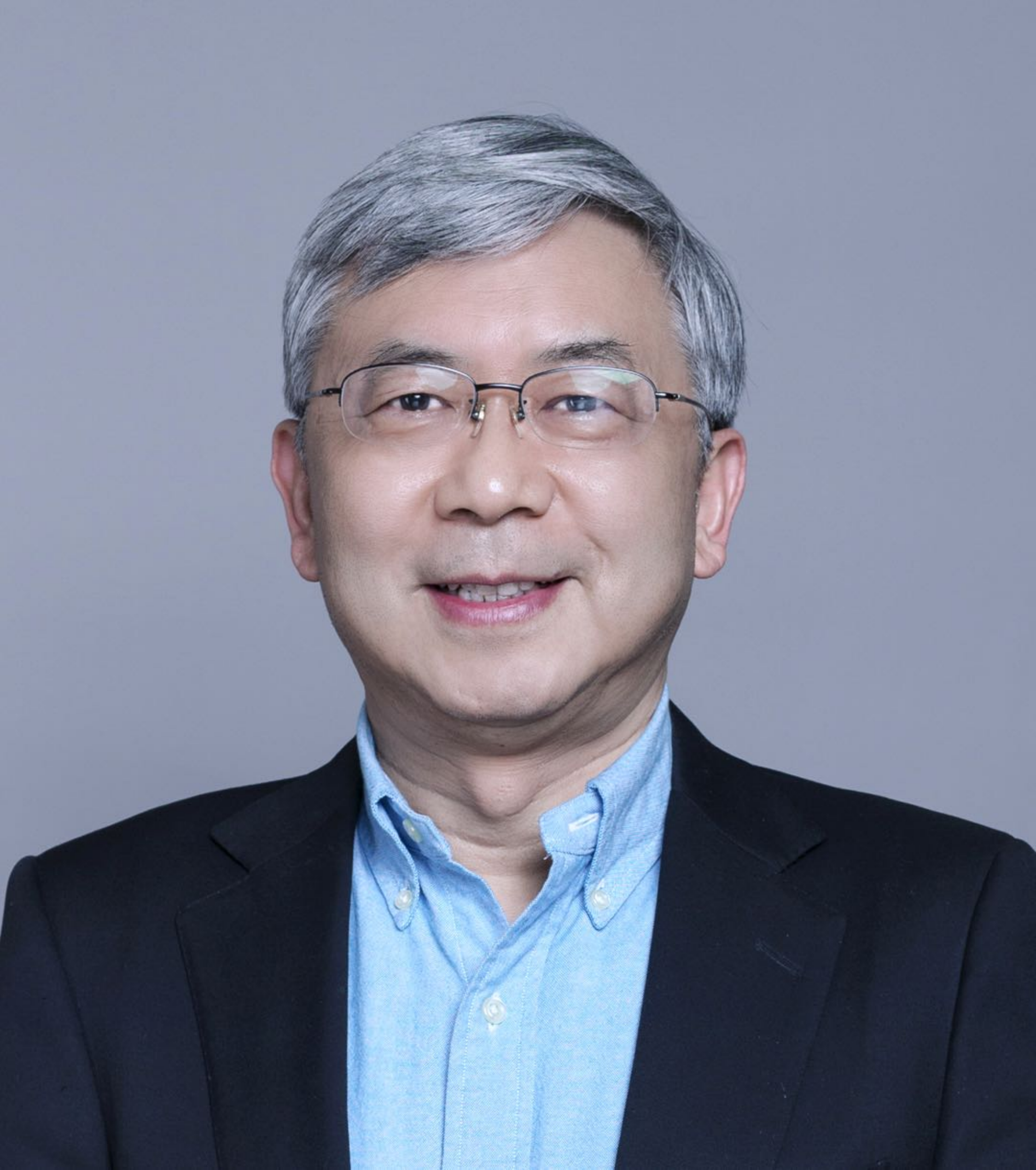}}]{Nanning Zheng}	(Fellow, IEEE) graduated in 1975 from the Department of Electrical Engineering, Xi'an Jiaotong  University, received the ME degree in Information and Control Engineering from Xi'an Jiaotong University in 1981, and a Ph.D. degree in Electrical Engineering from Keio University in 1985. He is currently a Professor and the Director of the Institute of Artificial Intelligence and Robotics at Xi'an Jiaotong University. His  research interests include computer vision, pattern recognition, computational intelligence, and hardware implementation of intelligent systems. He became a member of the Chinese Academy Engineering in 1999. He is a fellow of the IEEE.
\end{IEEEbiography}





\end{document}